\documentclass[conference]{IEEEtran}
\IEEEoverridecommandlockouts
\usepackage{caption}
\usepackage{subcaption}
\usepackage{cite}
\usepackage{amsmath,amssymb,amsfonts}
\usepackage{algorithmic}
\usepackage{graphicx}
\usepackage{textcomp}
\usepackage{xcolor}
\usepackage{multirow}
\usepackage[normalem]{ulem}
\useunder{\uline}{\ul}{}

\newcommand{\mm}{\operatorname{mm}} 
\newcommand{\nash}{\operatorname{nash}}
\newcommand{\cP}{{\mathcal P}}
\newcommand{\bs}{{\mathbf s}}
\newcommand{\ba}{{\mathbf a}}
\newcommand{\br}{{\mathbf r}}

\newcommand{\next}{\operatorname{next}}

\begin{document}

\title{Multi-agent Reinforcement Learning with Deep Networks for Diverse Q-Vectors}

\author{Zhenglong Luo, Zhiyong Chen, and James Welsh
        % <-this % stops a space
\thanks{The Authors are with the School of Engineering, The University of Newcastle, Callaghan, NSW 2308, Australia. Z. Chen is the corresponding author. E-mail: zhiyong.chen@newcastle.edu.au. }}

\maketitle

\begin{abstract}
Multi-agent reinforcement learning (MARL) has become a significant research topic due to its ability to facilitate learning in complex environments. In multi-agent tasks, the state-action value, commonly referred to as the Q-value, can vary among agents because of their individual rewards, resulting in a Q-vector. Determining an optimal policy is challenging, as it involves more than just maximizing a single Q-value. Various optimal policies, such as a Nash equilibrium, have been studied in this context. Algorithms like Nash Q-learning and Nash Actor-Critic have shown effectiveness in these scenarios. This paper extends this research by proposing a deep Q-networks (DQN) algorithm capable of learning various Q-vectors using Max, Nash, and Maximin strategies. The effectiveness of this approach is demonstrated in an environment where dual robotic arms collaborate to lift a pot.
\end{abstract}


\section{Introduction}

Traditional reinforcement learning (RL) algorithms typically focus on training a single agent to optimize its behavior in isolation. However, many real-world scenarios involve multiple agents that must learn to interact with each other and their environment. Thus, multi-agent reinforcement learning (MARL) is motivated by the need to develop intelligent systems capable of interacting and collaborating in complex environments. 

In the early literature of MARL, Hu and Wellman \cite{hu2003nash} extended Q-learning to a non-cooperative multi-agent setting by leveraging the framework of a generic stochastic game. In this approach, a learning agent maintains a Q-function representing joint actions and updates it based on Nash equilibrium behavior. Experimental validation was conducted using two distinct two-player grid games. A comparison of offline learning performance revealed that agents employing the Nash Q-learning method were more inclined to converge to the joint optimal path compared to those utilizing single-agent Q-learning. Notably, when both agents adopted Nash Q-learning, their performance exhibited a significant enhancement over instances where only single-agent Q-learning was employed.

This multi-agent Q-learning algorithm builds on the traditional Q-learning method \cite{watkins1992q}. By constructing virtual Q-tables, it effectively represents the Q-values corresponding to each action in each state. However, this approach has a significant limitation: the virtual Q-table expands proportionally with the growth of the state and action space. In MARL, this issue is exacerbated as the number of agents increases, leading to a substantial increase in the size of the virtual Q-table. Consequently, the algorithm is only practical for relatively small, discrete state and action spaces.
 
In light of this, the development of neural networks has led to the proposal of deep Q-networks (DQN) \cite{lecun2015deep} as a method to replace the Q-table with a neural network to approximate the Q-value. This methodology is applicable to large-scale, continuous state and action spaces. Additionally, deep learning neural networks utilize experience replay techniques, which allow networks to store previous experiences and derive training samples from these experiences. DQNs have been successfully applied in various fields, demonstrating their versatility and effectiveness \cite{van2016deep,silver2017mastering,wang2018deep}.

Building on the advantages of DQNs in handling large state and action spaces, researchers have attempted to integrate them with MARL. Some relevant works are discussed as follows. For example, the method proposed in \cite{tampuu2017multiagent} uses a neural network to approximate the Q-value function, enabling agents to learn effective policies in complex environments. However, this approximation often overlooks the influence of other agents on the environment. Although this approach simplifies modeling and enhances robustness, ignoring the policies and behaviors of other agents can lead to failure in finding optimal cooperation or competition strategies in complex interactive environments.

In \cite{foerster2016learning}, a unique communication channel for a multi-agent environment was designed, allowing agents to send and receive information over a shared communication channel. Each agent decides its next action based on the received communication information and its own state. Additionally, a joint training framework was proposed to optimize the policies of individual agents by maximizing the cumulative rewards of the team. The work in \cite{foerster2017stabilising} introduced important sampling weights and a goal network to stabilize the experience replay process. The stability of this approach was validated in cooperative navigation and cooperative communication tasks. These methods are based on DQN, with modifications to the communication methods and experience replay process.
 
Deep recurrent Q-learning (DRQN) and a meta-learning approach based on DQN were proposed in \cite{hausknecht2015deep} and \cite{al2017continuous}, respectively, to address challenges in multi-agent environments. DRQN excels in chase games by effectively handling partially observable Markov decision processes. The meta-learning algorithm, designed for non-smooth competitive multi-agent environments, enables agents to rapidly adapt to new tasks with only a few gradient updates.

The work in \cite{sutton1999policy} integrates DQN and policy gradient (PG) methods to promote cooperation in complex social dilemmas. The authors introduced an algorithm for multi-agent environments that uses independent DQNs and policy gradients, allowing each agent to independently learn and optimize its policy. In addition, the counterfactual multi-agent policy gradients (COMA) algorithm, proposed in \cite{foerster2018counterfactual}, evaluates each agent's contribution when sharing a global reward. Traditional PG techniques often suffer from high variance and low performance in multi-agent settings due to the interdependence of agents' rewards, with variance increasing exponentially with the number of agents. To mitigate this, deep deterministic policy gradient (DDPG) was adapted for multi-agent environments in \cite{lowe2017multi}. Furthermore, \cite{liu2021collaborative} combined Hindsight experience replay \cite{andrychowicz2017hindsight} and DDPG into the dual-arm deep deterministic policy gradient algorithm. This algorithm was validated in dual robotic arm simulation environments similar to those used in this paper, though our robotic arms operate based on real physical parameters.

The methods discussed above are based on DQN for multi-agent environments, but they are not directly designed for tasks requiring multi-agents to learn a joint optimal path, such as achieving a Nash equilibrium, as addressed in \cite{hu2003nash}. A direct extension of Nash Q-learning to DQN was proposed in \cite{casgrain2022deep}, demonstrating its feasibility in a complex game that studies the behavior of electronic exchanges. This approach employs two neural networks with different structures: the main neural network outputs the action and advantage function A, while the secondary neural network outputs the state value function approximation V. According to the paper, this structure resembles the actor-critic algorithm \cite{konda2000actor}, where the network outputting the action and A-value functions as the actor network, and the network outputting the V-value functions as the critic network. The Nash equilibrium is then solved using the output of the critic network to optimize the actor network. This motivates us to explore a direct DQN approach for multi-agent environments that retains the traditional DQN structure, rather than adopting an actor-critic framework.

Specifically, in this paper, we define various Q-vectors based on Max, Nash, and Maximin strategies, along with the corresponding optimal actions. To learn the optimal policies associated with each Q-vector, we develop a DQN-based algorithm. This algorithm retains the traditional DQN structure but replaces the maximization of a single Q-value with the optimization of a Q-vector. This modification more intuitively reflects the impact of incorporating game theory algorithms into traditional single-agent reinforcement learning algorithms in a multi-agent environment. The effectiveness of our algorithm is demonstrated in the Mujoco simulation environment using the task `two arms lift’ \cite{todorov2012mujoco}. It is noteworthy that such a dual robotic arm operation is rarely used to verify MARL algorithms due to its complexity. Instead, most MARL algorithms are tested in environments like tables, games, and mathematical models of electronic transactions. The physical robotic environment presents additional training challenges, such as friction, gravity, and the vibration of the robotic claw grip due to joint movement. Therefore, validation in such an environment highlights the reliability of our algorithm.

The remaining sections are organized as follows. Section~\ref{sec.Qvector} introduces various optimal Q-vectors based on Max, Nash, and Maximin strategies and formulates the MARL target. The DQN-based algorithm is discussed in Section~\ref{sec.Algorithm}. Section~\ref{sec.Experiments} presents experimental results from a robotic arm lifting a pot task to demonstrate the effectiveness of the algorithm. Finally, the paper is concluded in Section~\ref{sec.Conclusion}.

\section{Optimal Q-Vectors}

\label{sec.Qvector}

Consider $n\geq 2$ agents in a non-cooperative task. At each time instant $t$, the actions of the agents are denoted as $a_t^1, \ldots, a_t^n$, and the rewards are $r_t^1, \ldots, r_t^n$. The complete state of the multi-agent system is represented as $s_t$.  The behaviors of multiple agents are modeled as Markov decision processes with a state transition probability function $p(s_{t+1}|s_{t}, a_t^1, \ldots, a_t^n)$.
 
For example, in the task of two robotic manipulators lifting a pot, the reward for each robot is given by:
\begin{equation}
r_t^i = r_{{\rm height}, t} + r_{{\rm angle}, t} + r^i_{{\rm gripper}, t} + r^i_{{\rm action}, t}. \label{rti}
\end{equation}
Here, $r_{{\rm height}}$ represents the reward for the pot's height, which increases as the pot is lifted, and $r_{{\rm angle}}$ represents the reward for the pot's tilt angle, which decreases as the tilt angle increases. These rewards are shared by both robots.
Additionally, $r^i_{{\rm gripper}}$ represents the distance from the robot gripper to the pot's handle, where the reward increases as the distance decreases. Conversely, $r^i_{{\rm action}}$ accounts for the action cost, with a more negative value associated with larger actions.

 The future reward of agent $i$, also known as the return, is defined as:
\begin{align}
R^i_t = \sum_{k=0}^\infty \gamma^k r^i_{t+k+1}.
\end{align}
Here, the discounting factor $\gamma \in [0, 1]$ penalizes rewards in the future. The policy for agent $i$ is denoted as $\pi^i(a_{t}^i | s_t)$, which generates $a_{t}^i$ based on the current state $s_t$.
For the convenience of presentation, denote 
$a=(a^1,\cdots, a^n)$, $r=(r^1,\cdots, r^n)$, and
 $\pi=(\pi^1,\cdots, \pi^n)$.  
The Q-value of the state-action pair for agent $i$, following the policy $\pi$, is defined as:
\begin{align}
Q^i_\pi (s,a) = {\mathbb E}_\pi [R^i_t  |  s_t=s, a_t=a].
\end{align}
Denote $Q=(Q^1,\cdots, Q^n)$.

In this multi-environment context, the Q-vector cannot be defined based solely on a single maximum return. Instead, we consider three different definitions of optimal Q-vectors, denoted as $Q^* = (Q^{*1},\cdots, Q^{*n})$ in this paper. 

\begin{itemize}
\item  Max Q-vector:

The max Q-vector is based on each agent's individual maximum policy. For each agent $i$, we define
\begin{align*}
\pi_{\max[i]}=(\pi^1_{\max[i]},\cdots, \pi^n_{\max[i]}) = \arg\max_\pi Q^i_\pi(s, a).
\end{align*}
The policy $\pi_{\max[i]}$ is designed to maximize the Q-value for agent $i$, but it cannot be directly implemented because, in general, 
$\pi_{\max[i]} (a_{t} | s_t) \neq \pi_{\max[j]} (a_{t} | s_t)$ for $i \neq j$.
However, one way to implement the policy $\pi_{\max[i]}$ is to have it provide actions only for agent $i$, that is, $\pi^i_{\max[i]} (a^i_{t} | s_t)$. Thus, it forms a combined policy as follows:
\begin{align*}
\pi^*_{\max}=(\pi^1_{\max[1]}, \cdots, \pi^n_{\max[n]}).
\end{align*}
Based on this policy, the corresponding Q-vector is the max Q-vector, that is,
\begin{align*}
Q^*_{\max}(s, a) = Q_{\pi_{\max}}(s, a).
\end{align*}

\item Nash Q-vector:

A Nash policy is one where each agent chooses their optimal policy given the policies chosen by other agents, such that no agent can benefit by unilaterally changing their own policy.  A joint multi-agent Nash policy $\pi^*_{\nash}$ is defined as a policy satisfying:
 \begin{align*}
Q^i_{\pi^*_{\nash}}(s,a)  \geq  \nonumber Q^i_{(\pi^i, \pi^{*-i}_{\nash})}(s,a),\; \forall \pi^i , \forall i. 
\end{align*}
Here, the notation $\pi^{-i}$ denotes the polices taken by all other agents, that is,  
 \begin{align*}
 \pi^{-i} = (\pi^{1}, \cdots, \pi^{i-1},  \pi^{i+1}, \cdots,  \pi^{n}). 
 \end{align*}
The corresponding Nash Q-vector is defined when the agents follow the joint Nash policy $\pi^*_{\nash}$, as follows:
 \begin{align*}
Q^{*}_{\nash} (s,a) = Q_{\pi^*_{\nash}}(s,a).  
\end{align*}

\item Maximin Q-vector:

The maximin Q-value of an agent is the highest Q-value that an agent can be sure to obtain without knowing the policies of the other agents, and the corresponding policy is, 
\begin{align}
 \pi^{*i}_{\mm}=\arg\max_{\pi^i} \min_{\pi^{-i}}
    Q^i_{(\pi^i, \pi^{-i})} {(s,a)}.   
\end{align}
It forms a joint maximin policy $\pi^*_{\mm}=(\pi^{*1}_{\mm}, \cdots, \pi^{*n}_{\mm})$, and the corresponding maximin Q-vector is
   \begin{align*}
Q^{*}_{\mm} (s,a) = Q_{\pi^*_{\mm}}(s,a).  
\end{align*}

\end{itemize}

The primary objective is to develop reinforcement learning strategies using deep neural networks to learn the optimal policies associated with each optimal Q-vector. The state-action value function is estimated by a neural network parameterized by $\phi$ and the value is denoted as $Q_{\phi}(s,a)$. 
In the conventional single-agent scenario, the optimal action $a$ can be found by selecting the action with the maximum Q-value, defined as: $a^*=  \arg\max_a \{ Q_\phi(s, a) \}$.  
In the multi-agent case, the selection of action based on the Q-vector is more complicated, denoted as:
\begin{align}
a^*=  \cP \{ Q_\phi(s, a) \}. \label{cP}
\end{align}
The specific definition of the operator $\cP$ is provided below, along with the different definitions of Q-vectors.

\begin{itemize}

\item Max Q-vector ($\cP_{\max}$): the operator $\cP$ is denoted as $\cP_{\max}$, which maps $Q_\phi(s, a)$ to $a^*$, denoted as $a^*_{\max}$, according to:
\begin{align*}
a^*_{\max}=&(a^1_{\max[1]}, \cdots, a^n_{\max[n]}) \\
a_{\max[i]}=&(a^1_{\max[i]},\cdots, a^n_{\max[i]}) = \arg\max_a Q^i_\phi(s, a).
\end{align*}

\item Nash Q-vector ($\cP_{\nash}$): the operator $\cP$ is denoted as $\cP_{\nash}$, which maps $Q_\phi(s, a)$ to $a^*$, denoted as $a^*_{\nash}$, according to:
\begin{align*}
Q^i_{\phi}(s,a^*_{\nash})  \geq  \nonumber Q^i_{\phi}(s,(a^i, a^{*-i}_{\nash})),\; \forall a^i , \forall i. 
\end{align*}
 
\item Maximin Q-vector ($\cP_{\mm}$):  the operator $\cP$ is denoted as $\cP_{\mm}$, which maps $Q_\phi(s, a)$ to $a^*$, denoted as $a^*_{\mm}$, according to:
\begin{align*}
a_{\mm}^*=&(a^{*1}_{\mm},\cdots, a_{\mm}^{*n})\\
 a^{*i}_{\mm}=&\arg\max_{a^i} \min_{a^{-i}}
    Q^i_\phi {(s,(a^i, a^{-i}))}.   
\end{align*}

 \end{itemize}

With the Q-vectors and the corresponding action selection operator $\cP$ defined, we are ready to introduce the multi-agent reinforcement learning algorithms.

\section{Algorithms}\label{Multi}

\label{sec.Algorithm}

A DQN framework is used to develop the reinforcement learning algorithms based on Q-vectors. The learning process using DQN is composed of three main components:

\begin{itemize}
\item Experience reply buffer $B$;

\item Prediction networks $Q_\phi(s,a)$, parameterized by $\phi$;

\item Target networks $Q_\theta(s,a)$, parameterized by $\theta$. 
\end{itemize}

In a multi-agent scenario, both $Q_\phi(s,a)$ and $Q_\theta(s,a)$ generate Q-vectors. These Q-vectors consist of $n$ neural networks, one for each agent.
The learning algorithm is discussed below.

\begin{enumerate}
 
\item Measure the state $s_t$ and input it into the prediction network to obtain the Q-vector for all possible actions in state $s_t$. This Q-vector is denoted as $Q_{\phi_t}(s_t, a)$, where $\phi_t$ represents the parameters of the prediction network at time $t$.
 
\item Select the optimal action according to \eqref{cP}:
\begin{align}
a^* =  \cP \{ Q_{\phi_t}(s_t, a) \}.
\end{align}
Noting $\cP$ has different operations corresponding to the definition of the Q-vector.
Then, choose the action using the epsilon-greedy policy. With a probability of $\epsilon$, select a random action $a_t$, and with a probability of $1-\epsilon$, choose $a_t = a^*$.

\item Perform the action $a_t$ and move to a new state $s_{t+1}$. Note that $a_t$ is an action vector consisting of the actions of all agents.

\item  Record the action $a_t$ and the reward $r_{t+1}$ from all agents and update the experience replay buffer as $\{s_t, s_{t+1}, a_t, r_{t+1}\} \rightarrow B$.

\item  Sample some random batches $\{\bs, \bs_{\next}, \ba, \br\}$ of transitions from the replay buffer and calculate the target using the target network:
\begin{align} 
\br + \gamma {Q}_{\theta_t} (\bs_{\next},a^*)   \label{target}
\end{align} 
Here, $\theta_t$ represents the parameters of the target network at time $t$. The optimal action is found using the prediction network as:
\begin{align*}
a^*=  \cP \{ Q_{\phi_t}(\bs_{\next}, a) \}.  
\end{align*}

\item According to 
\begin{align*}
Q^*(s_t,a_t) = \mathbb{E}[r_{t+1}] + \gamma Q^*(s_{t+1}, \pi^*(a_{t+1} | s_t)),
\end{align*}
the target in \eqref{target} is used for training the prediction network by minimizing the loss defined as:
\begin{align*}
L &= \left\{ Q_{\phi_t} (\bs, \ba) - [\br + \gamma Q_{\theta_t} (\bs_{\next}, a^*)] \right\}^2.
\end{align*}
In particular, gradient descent is performed with respect to the prediction network parameters to minimize this loss. Consequently, the network parameters are updated from $\phi_t$ to $\phi_{t+1}$.

\item  If $t \neq 0 \mod C$ (i.e., $t$ is not a multiple of $C$), keep the target network unchanged: $\theta_{t+1} =\theta_t$. Otherwise, if $t =0 \mod C$ (i.e., after every $C$ iterations), copy the prediction network weights to the target network weights: $\theta_{t+1} =\phi_{t+1}$.
 
\end{enumerate}

Repeat these steps for $M$ episodes.

\medskip

In a multi-agent scenario, both $Q_\phi(s,a)$ and $Q_\theta(s,a)$ generate Q-vectors. These Q-vectors consist of $n$ neural networks, one for each agent. Specifically, each agent $i$ can maintain a pair of neural networks, $Q^i_\phi(s,a)$ and $Q^i_\theta(s,a)$, and exchange Q-values with other agents to implement the algorithm. Alternatively, each agent can maintain a complete copy of the prediction and target networks, $Q_\phi(s,a)$ and $Q_\theta(s,a)$, independently, without needing to request Q-values from other agents.

The above Q-vector based learning algorithm is introduced in the basic DQN framework. It can also be implemented with variants of DQN frameworks. For instance, it can be implemented with the dueling DQN (DDQN) algorithm, where a neural network produces the state-value $V(s)$ and the advantage function $A(s,a)$, which together determine the state-action value $Q(s,a)$. This DDQN algorithm is used in our experiments.

\section{Experiments}
\label{sec.Experiments}

To evaluate the effectiveness of Q-vector based algorithms with various definitions of Q-vector, we conducted experiments using the Robosuite environment \cite{zhu2020robosuite}. Robosuite is a modular simulation framework built on the MUJOCO physics engine \cite{todorov2012mujoco}, providing several pre-built simulation environments such as `2 arms lift a pot' and `block lifting'. Additionally, the authors have developed a robot arm model based on real structures, ensuring accurate representation of the robot arm's size and dynamic performance of its individual joints and grippers, thus establishing it as our primary simulation environment.

\subsection{Experimental Environment}

\begin{figure}
\centering
\includegraphics[width=0.2\textwidth]{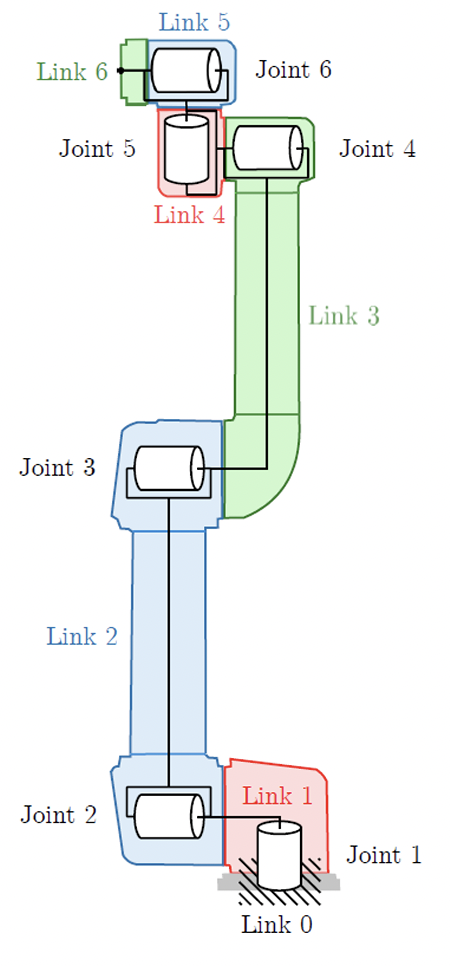}
\caption{Schematic diagram of UR5e joints.}
\label{Joint}
\end{figure}

To align the simulation with the real robotic arm system in the lab, the UR5e model was chosen to represent the robotic arms in the environment, with its joints illustrated in Figure~\ref{Joint}. The grippers were left at their default settings. The parameter {\it `envconfiguration`} was set to {\it `single'} to position the two robotic arms on opposite sides of the table. Additionally, {\it `has renderer'} was set to {\it `True'} to enable real-time observation of the robotic arm movements.  The {\it `control frequency`} parameter, which determines the frequency of the input signal, was set to {\it `control frequency=4'}. The {\it `horizon`} parameter, defining the end of the current cycle after a specified number of actions have been executed and the environment initialized, was set to {\it `horizon=100'}. These settings allowed the two robotic arms to raise the pot to its highest position while following the ideal path, thereby ensuring effective training completion.

\begin{figure}
    \centering
    \includegraphics[width=0.45\textwidth]{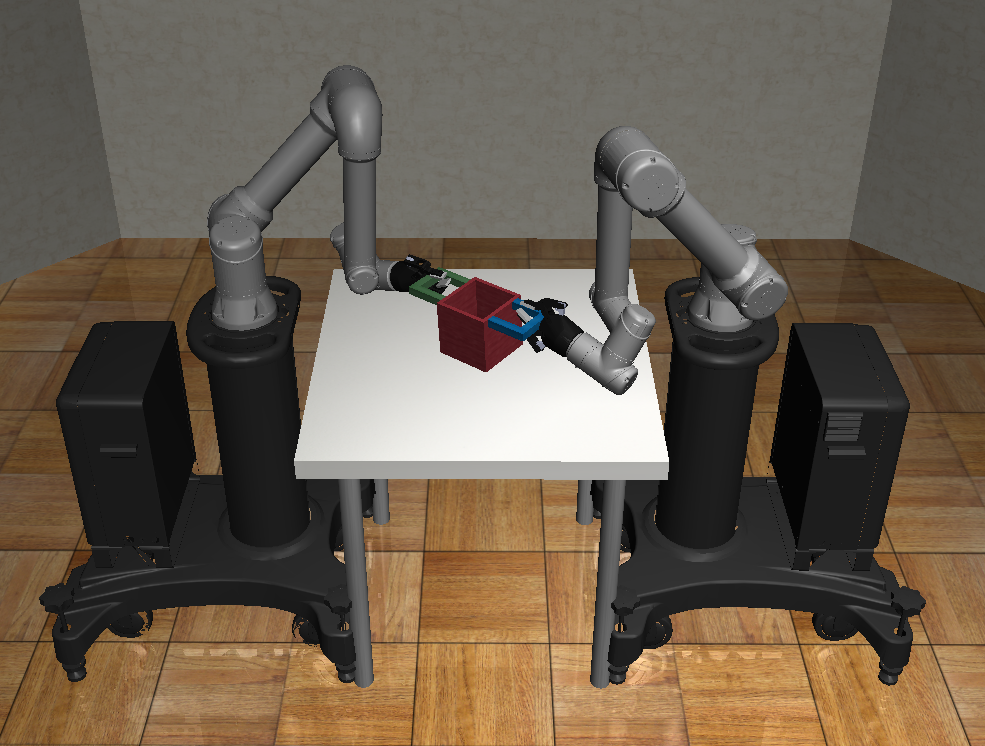}
    \caption{Initial position of the robot arms lifting task.}
    \label{originlposition}
%\end{figure}
\medskip
%\begin{figure}
    \centering
    \includegraphics[width=0.45\textwidth]{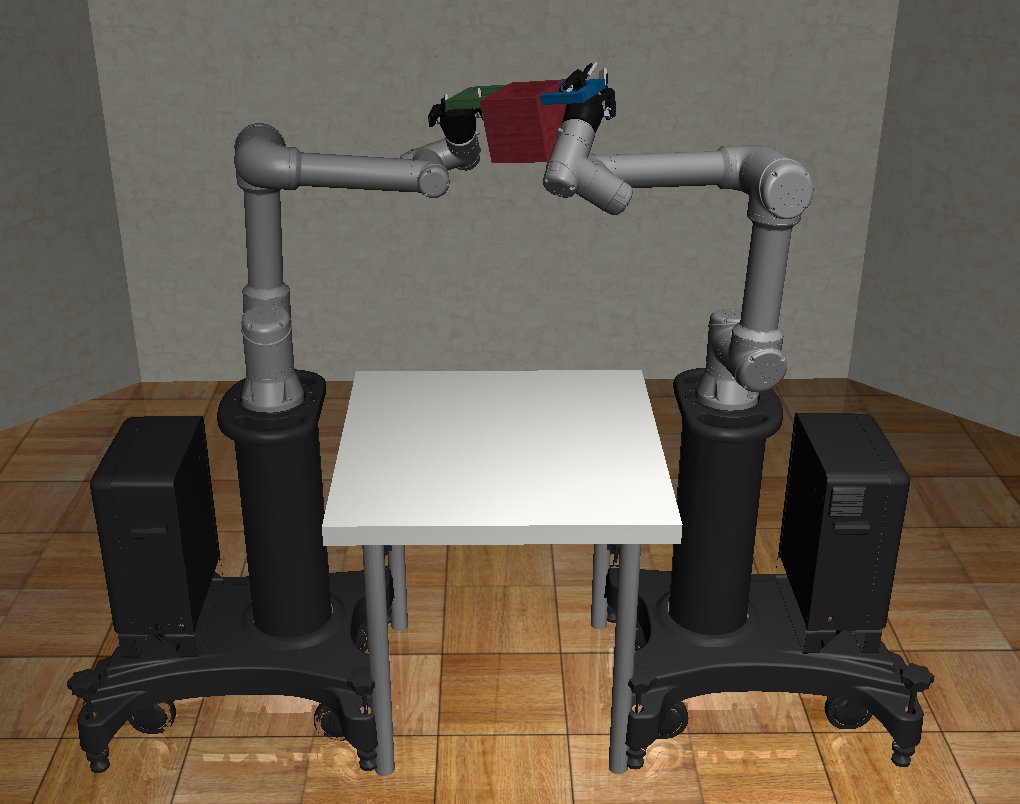}
    \caption{Final position of the robot arms lifting task.}
    \label{finalposition}
\end{figure}

In the ‘2 arms lift a pot’ Robosuite environment, two robotic arms were tested in the experiments, creating a multi-agent environment. For convenience of presentation, the first arm with $i=1$ is referred to as the left arm, and the second arm with $i=2$ is referred to as the right arm. The main task is for the two robotic arms to lift the centrally located pot simultaneously by controlling the joints and keeping the pot as smooth as possible in the process. The state $s_t\in {\mathbb R}^{12}$ consists of the angles of six joints of both robotic arms. The initial position is shown in Figure \ref{originlposition}, and the position when the pot is lifted is depicted in Figure \ref{finalposition}.

The variable {\it `Joint Velocity'}, provided by Robosuite, is used as the control action. Specifically, we exercised actions on two joints of each robotic arm, i.e., $a_t^i = (a_t^{i1}, a_t^{i2}) \in {\mathbb R}^2$, rather than all six joints, where $a_t^{i1}$ is Joint 2 and $a_t^{i2}$ is Joint 3 in Figure~\ref{Joint}. This deliberate choice guarantees the attainment of the experimental objective: evaluating the training efficiency and facilitating the observation of performance disparities among various algorithms. In this manner, the movement of each joint is discretized into two distinct actions: one involving elevation and the other involving immobility, that is, $a_t^{i1}, a_t^{i2} \in \{-0.1, 0\}$.

The reward $r_i^t$ is defined as \eqref{rti}. There is no action cost associated with Joint 3. Let the action cost of Joint 2 be $c^i$. More specifically, $r^i_{{\rm action}} = 0$ or $c^i$, when $a_t^{i1} = 0$ or $-0.1$, respectively. We set two types of action costs to evaluate the algorithms: $c = (c^1, c^2) = (-5, -5)$ and $c = (0, -5)$. In the first case, the action of each agent's Joint 2 has the same cost of $-5$. In the second case, the action of Joint 2 of the left arm is free of cost, while that of the right arm has a cost of $-5$.

\subsection{Results and Evaluation}

For the two joints of the arms, the action $a_t\in {\mathbb R}^4$, with each joint action taking two possible actions: elevation ($-0.1$) and immobility ($0$). This results in 16 combinations of actions. Since the two joints of each arm contribute to the robot's behavior in a similar manner, we combine them as a virtual action for simplicity of explanation. Essentially, each arm has two actions: elevation ($-0.1$) and immobility ($0$), leading to four combinations of actions: $(0, 0)$, $(-0.1, 0)$, $(0 -0.1)$, and $(-0.1, -0.1)$. We can establish Q-vector variation tables, which indicate the corresponding actions based on the optimal policies, i.e., Max, Nash, and Maximin. These optimal policies reflect the corresponding behaviors of the two arms. It is expected that the proposed learning algorithms can successfully learn these behaviors in all scenarios.

Below, we examine two cases: one for $c = (-5, -5)$ and another for $c = (0, -5)$.

\subsubsection{Case 1: Balanced Action Costs}

When one arm acts to elevate while the other remains immobile, their contribution to the height of the pot results in a reward increase denoted by $p_1 > 0$.
When both arms act to elevate, they jointly contribute to the height of the pot, resulting in a reward increase denoted by $p_2 > p_1$. With the reward weights set accordingly, it is reasonable to assume $p_1 > p_2 - 5$ and $p_1 > 5$.  The balanced cost of action $c = (-5, -5)$ is considered here, meaning that the elevation action of each arm incurs an equal cost of $-5$.

\begin{figure*}[t]
\begin{minipage}{0.48\textwidth}
\centering
\includegraphics[width=1\textwidth]{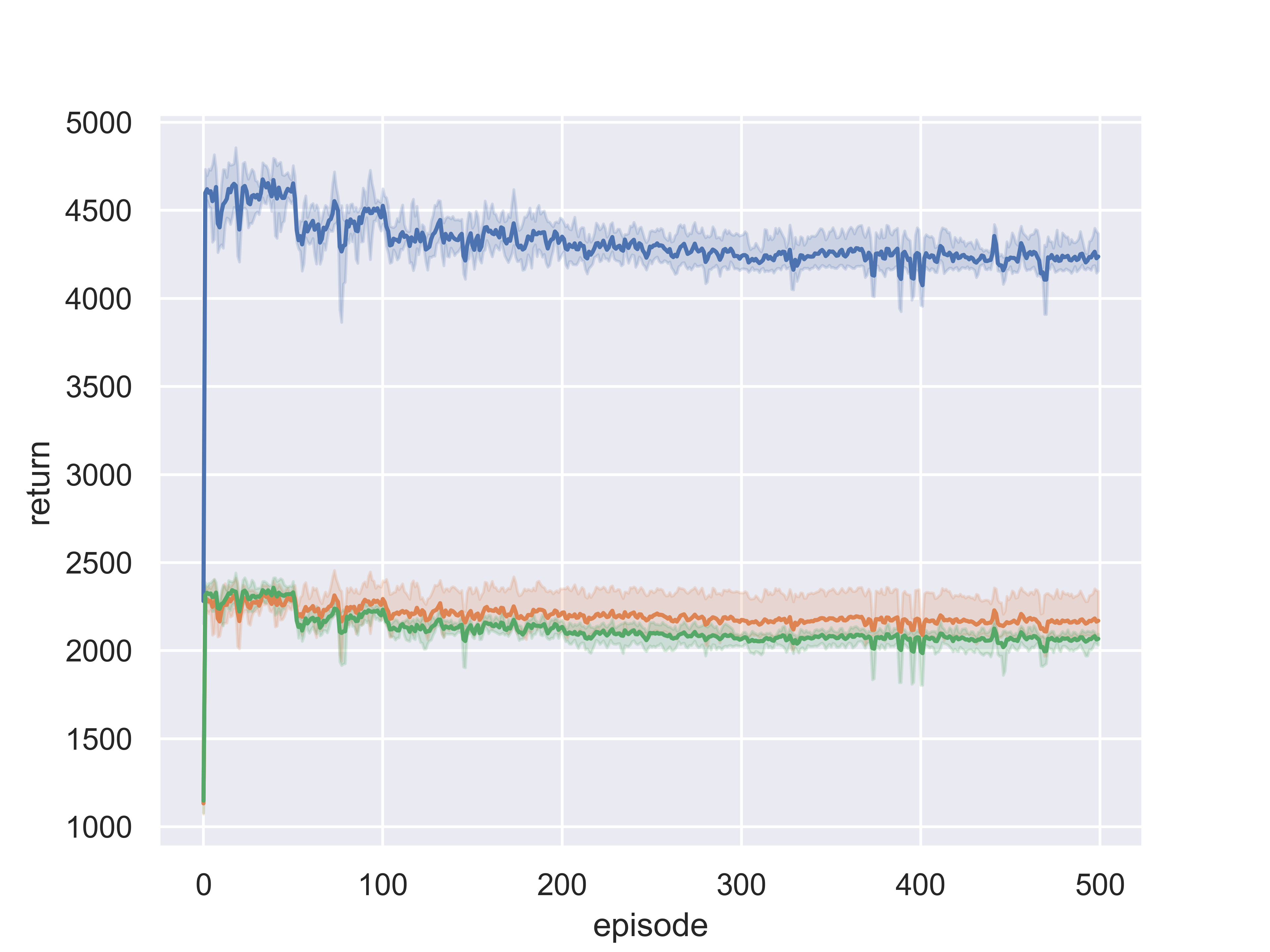}
\caption{Profile of Max Q-vectors in Case 1: no arm lifted.}
\label{Max55} 
\end{minipage} \hspace{0.03\textwidth}
\begin{minipage}{0.48\textwidth}
\centering
\includegraphics[width=1\textwidth]{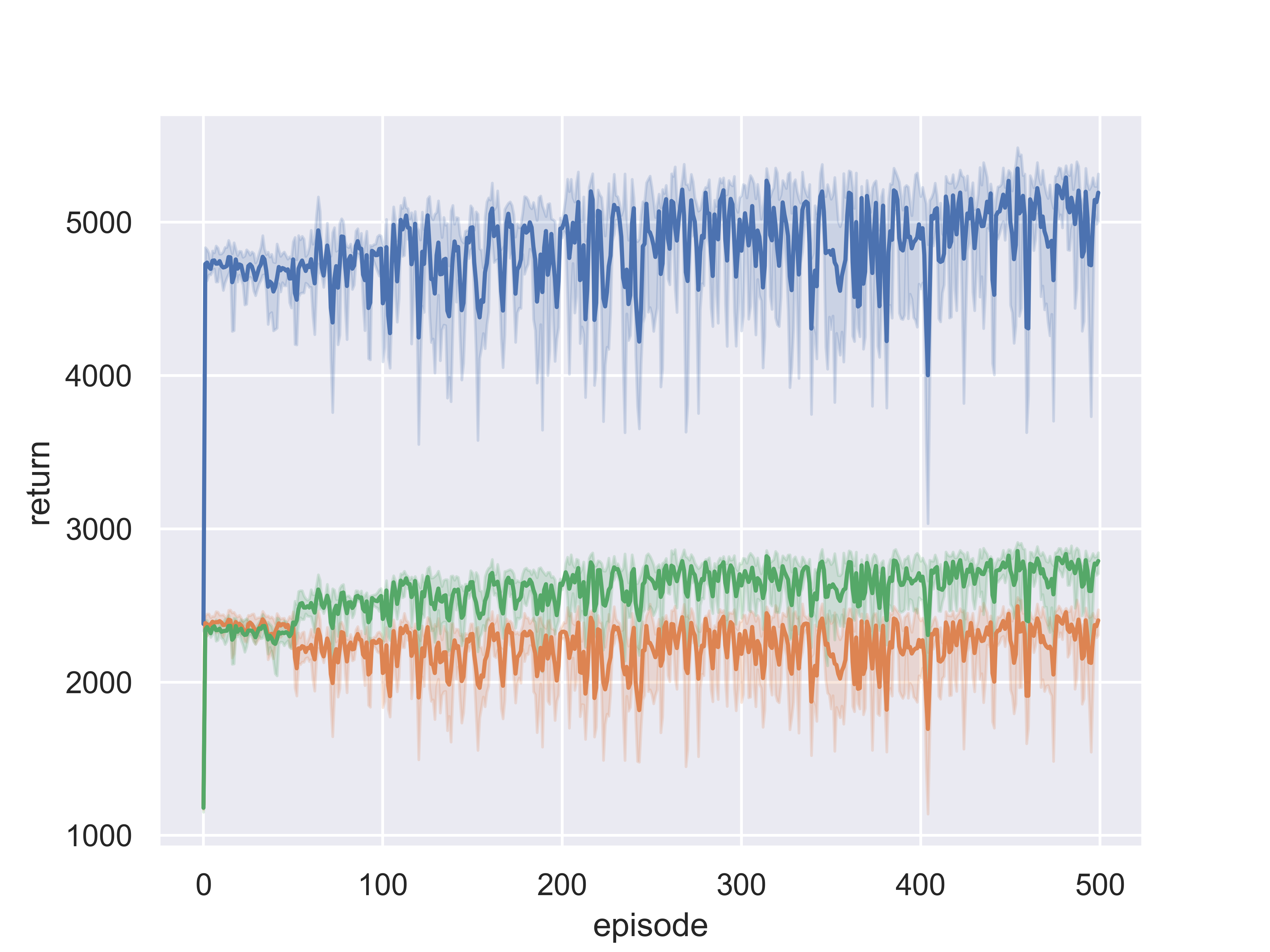}
\caption{Profile of Nash Q-vectors in Case 1: right arm lifted.}
\label{Nash55}
\end{minipage} \hspace{0.03\textwidth}
\begin{minipage}{0.48\textwidth}
\centering
\includegraphics[width=1\textwidth]{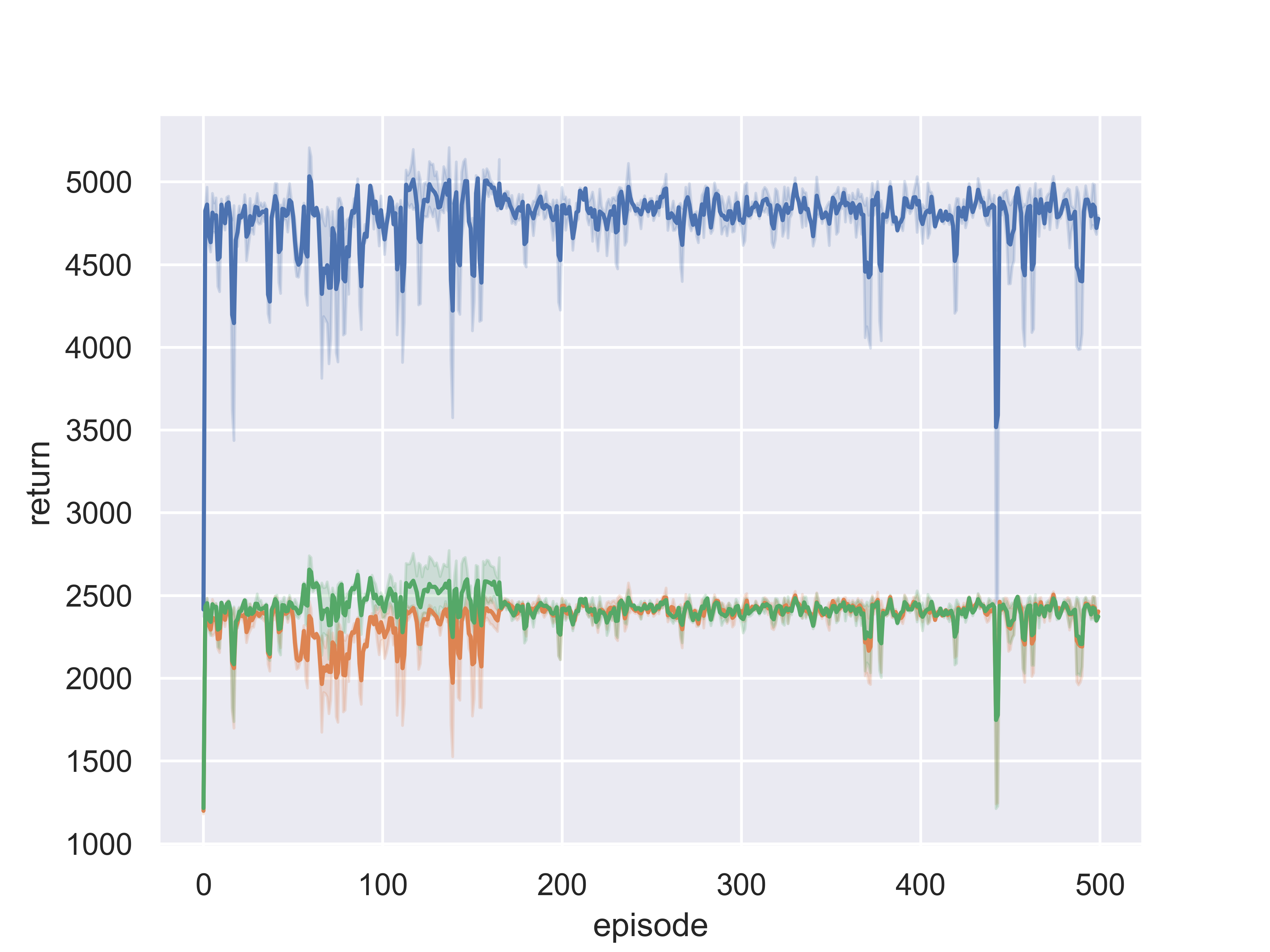}
\caption{Profile of Nash Q-vectors in Case 1:  right arm lifted and then left arm lifted.}
\label{Nash55_1}
\end{minipage} \hspace{0.03\textwidth}
\begin{minipage}{0.48\textwidth}
\centering
\includegraphics[width=1\textwidth]{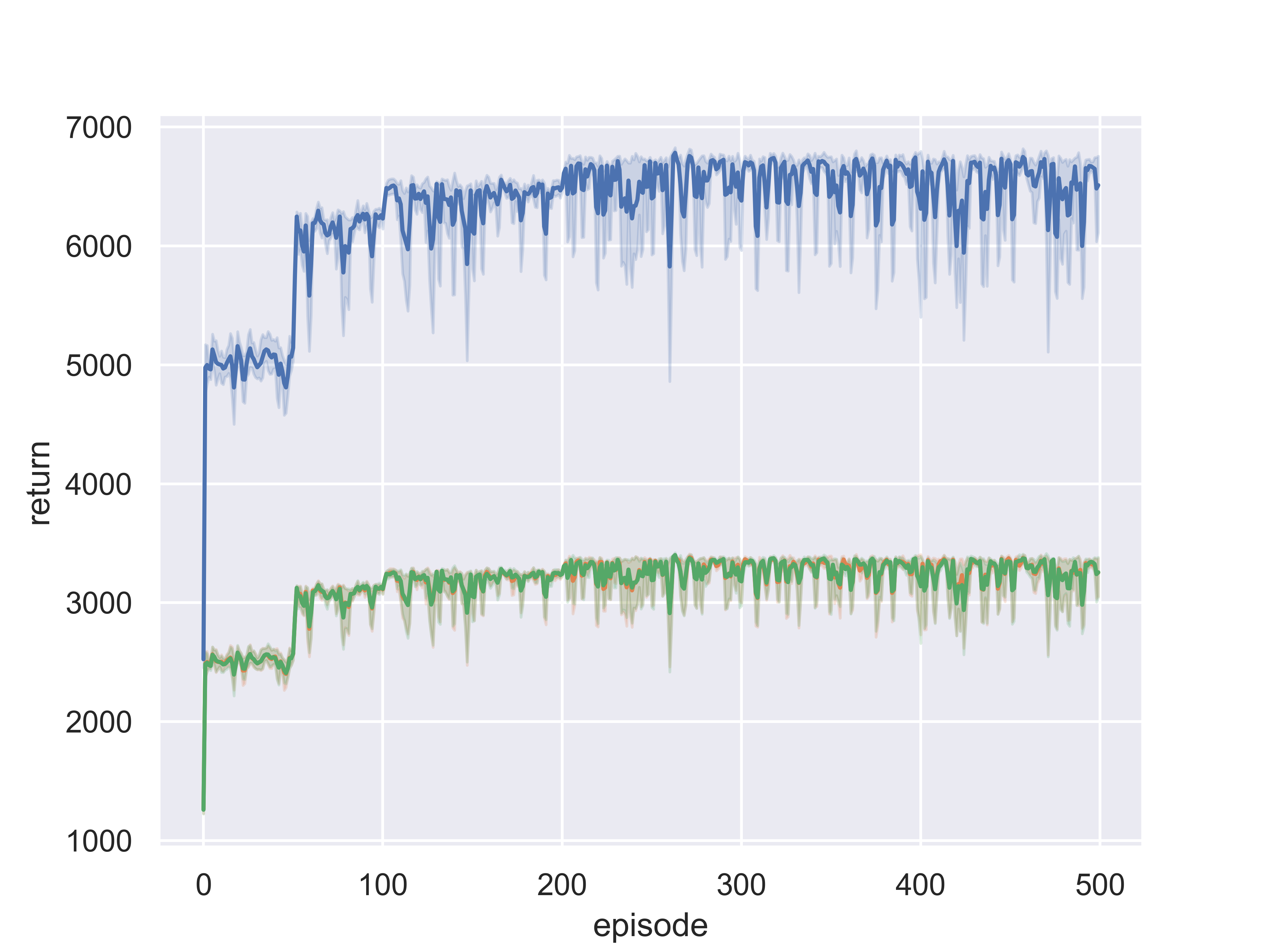}
\caption{Profile of Maximin Q-vectors in Case 1: two arms lifted.}
\label{MM55}
\end{minipage} \hspace{0.04\textwidth} 
\end{figure*}

The Q-vector variations corresponding to three different types of Q-vectors are then listed in Table~\ref{tablecase1}. The experimental results are presented in Figures~\ref{Max55} through~\ref{MM55}. In the figures, the return of the left arm is shown in orange, the return of the right arm is depicted in green, and their total is represented in blue.
Each experiment was repeated six times.

\begin{table}[h]
\centering
\begin{tabular}{|cc|cc|}
\hline
\multicolumn{2}{|c|}{\multirow{2}{*}{\bf Max: $\mathbf{(0, 0)}$}} & \multicolumn{2}{c|}{Right arm}  \\ 
  & & 0 & -0.1 \\
\hline \multirow{2}{*}{Left arm} & 0 & $(0,\; 0)$ &  $(\mathbf{p_1},\; p_1-5)$ \\
& -0.1 & $(p_1-5,\; \mathbf{p_1})$ &  $(p_2-5,\; p_2-5)$ \\
\hline
\hline
%%%
\multicolumn{2}{|c|}{\bf Nash: $\mathbf{(-0.1, 0)}$ } & \multicolumn{2}{c|}{Right arm}  \\ 
\multicolumn{2}{|c|}{or $\mathbf{(0, -0.1)}$}  & 0 & -0.1 \\
\hline \multirow{2}{*}{Left arm} & 0 & $(0,\; 0)$ &  $\mathbf{(p_1,\; p_1-5)}$ \\
& -0.1 & $\mathbf{(p_1-5,\; p_1)}$ &  $(p_2-5,\; p_2-5)$ \\
\hline
\hline
%%%
\multicolumn{2}{|c|}{\bf Maximin:} & \multicolumn{2}{c|}{Right arm}  \\ 
\multicolumn{2}{|c|}{ $\mathbf{(-0.1, -0.1)}$} & 0 & -0.1 \\
\hline \multirow{2}{*}{Left arm} & 0 & $(0,\; 0)$ &  $(p_1,\; \mathbf{p_1-5})$ \\
& -0.1 & $(\mathbf{p_1-5},\; p_1)$ &  $(p_2-5,\; p_2-5)$ \\
\hline
\hline
\end{tabular}
\vspace{3mm}
\caption{Q-vector variations and optimal actions in Case 1.}
\label{tablecase1}
\end{table}
%$p_2>p_1 >p_2-5$, $p_1>5$

First, for the Max Q-vector, it is shown in Table~\ref{tablecase1} that the optimal action is $(0,0)$. The return for this case is depicted in Figure~\ref{Max55}. With the learned optimal policy, the arms remained stationary in the initial position or executed only minor movements, without lifting either arm. This outcome aligns with the expected behavior of the optimal action $(0,0)$. It is worth mentioning that in one experiment, one arm was lifted, which was considered an unsuccessful learning instance.

Secondly, for the Nash Q-vector, Table~\ref{tablecase1} shows that the optimal action is $(-0.1,0)$ or $(0, -0.1)$. Our algorithms randomly select one of these multiple optimal actions. The experimental results are presented in Figures~\ref{Nash55} and \ref{Nash55_1}. In Figure~\ref{Nash55}, which includes four experiments, it is observed that the right arm is lifted with the learned policy, aligning with the optimal action $(0, -0.1)$. Figure~\ref{Nash55_1} shows that the tilt angle of the pot was excessively steep due to one arm being lifted. Consequently, one robot arm would cease motion until the other arm continued to move, thereby reducing the tilt angle before resuming action. This behavior aligns with the existence of two optimal actions.

Thirdly, for the Maximin Q-vector, the optimal action is $(-0.1, -0.1)$, which implies that both arms are lifted. This objective was successfully accomplished by the learned policy, as demonstrated in Figure~\ref{MM55}. In this case, the pot was lifted as shown in Figure~\ref{finalposition}. These experiments verify that all the Q-vector-based optimal policies can be successfully learned by the proposed Dueling DQN algorithm in Case 1.

\begin{figure*}[t]
\begin{minipage}{0.48\textwidth}
\centering
\includegraphics[width=1\textwidth]{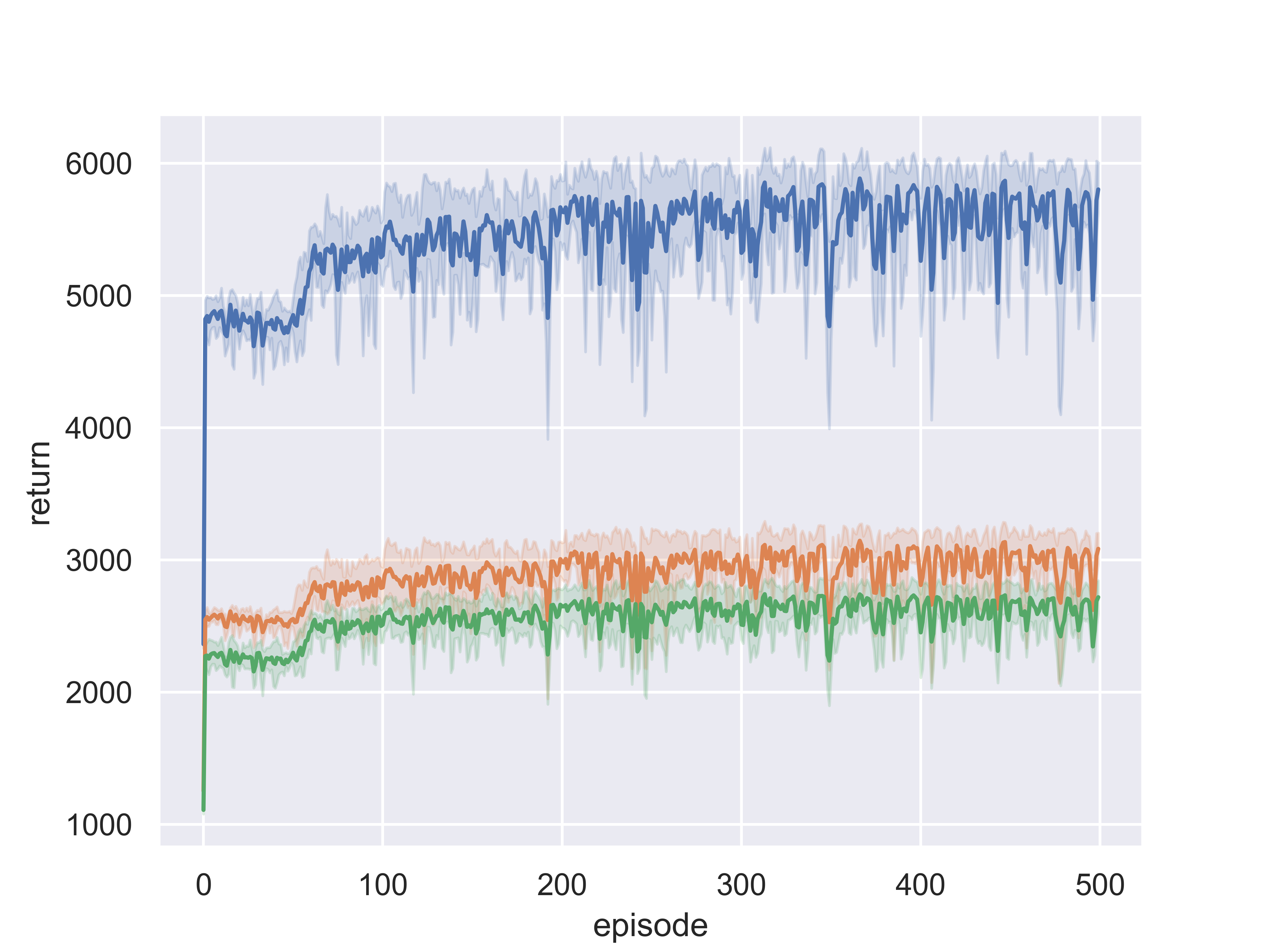}
\caption{Profile of Max Q-vectors in Case 2: left arm lifted.}
\label{Max05}
\end{minipage} \hspace{0.03\textwidth}
\begin{minipage}{0.48\textwidth}
%\end{figure}
%\begin{figure}
\centering
\includegraphics[width=1\textwidth]{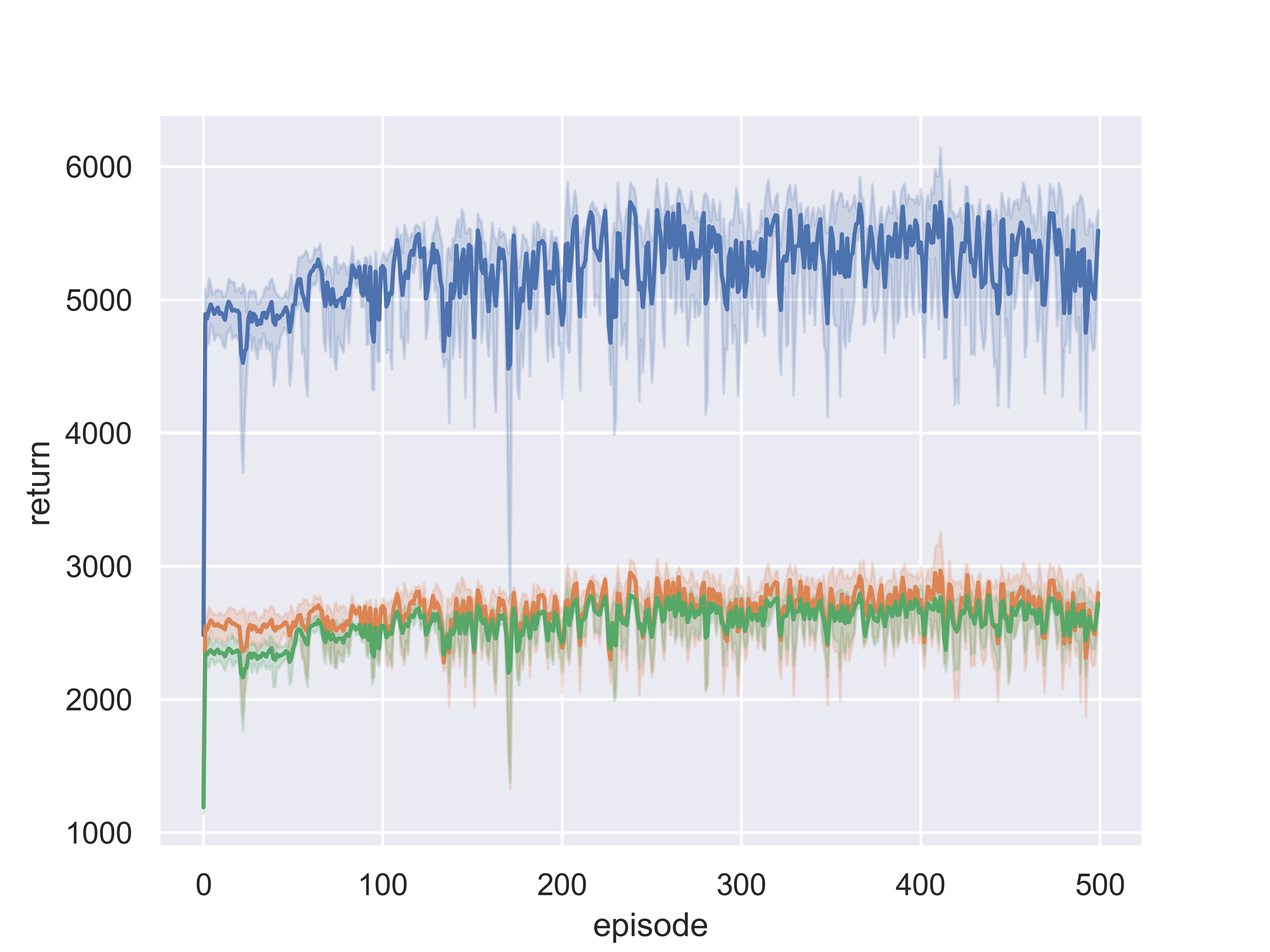}
\caption{Profile of Nash Q-vectors in Case 2: left arm lifted.}
\label{Nash05} 
\end{minipage}
\end{figure*}
\begin{figure*}[t]
\begin{minipage}{0.48\textwidth}
\centering
\includegraphics[width=1\textwidth]{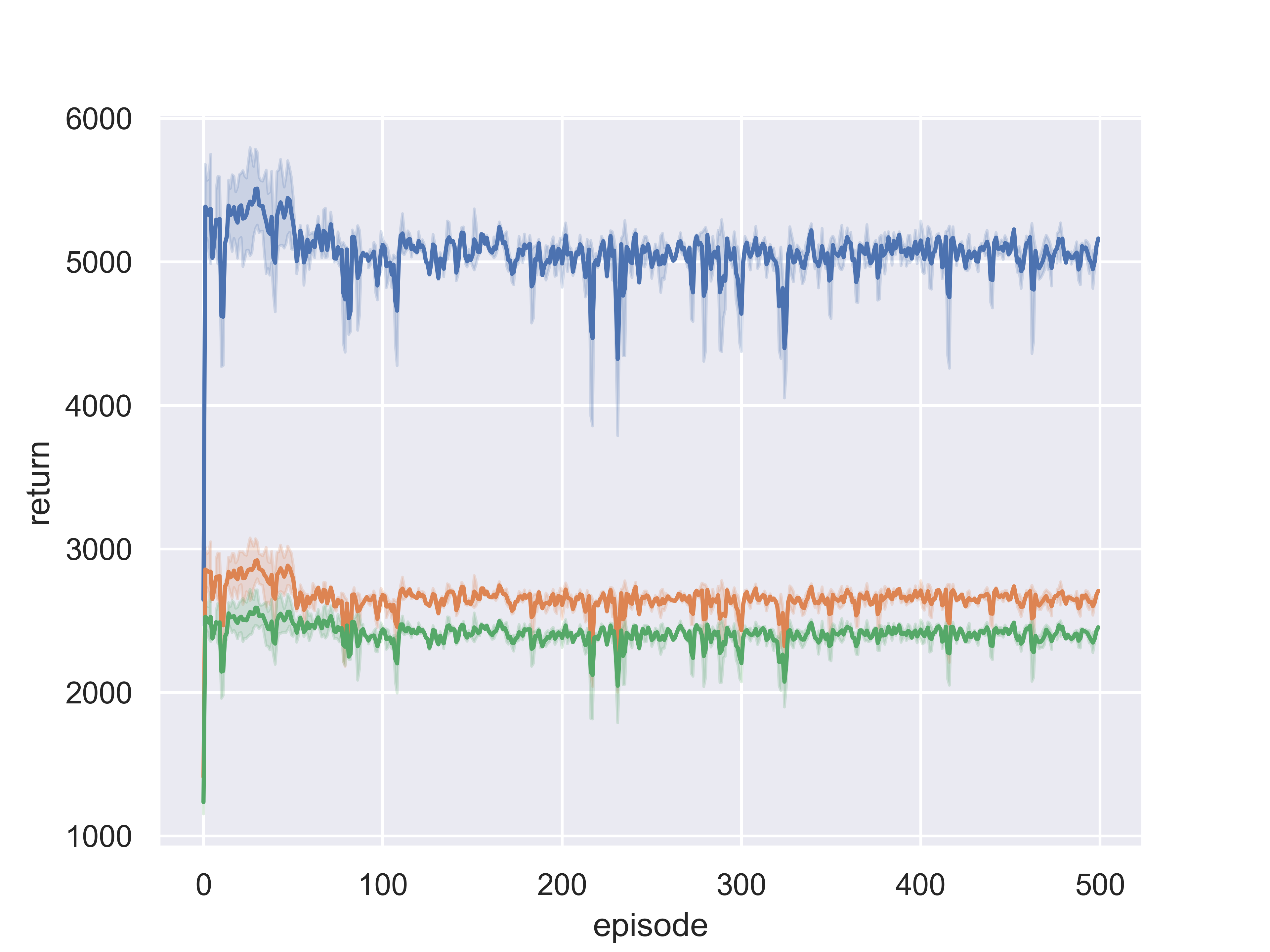}
\caption{Profile of Nash Q-vectors in Case 2:  left arm lifted and then right arm lifted.}
\label{Nash05_1}
\end{minipage} \hspace{0.03\textwidth}
%\end{figure}
%\begin{figure}
\begin{minipage}{0.48\textwidth}
\centering
\includegraphics[width=1\textwidth]{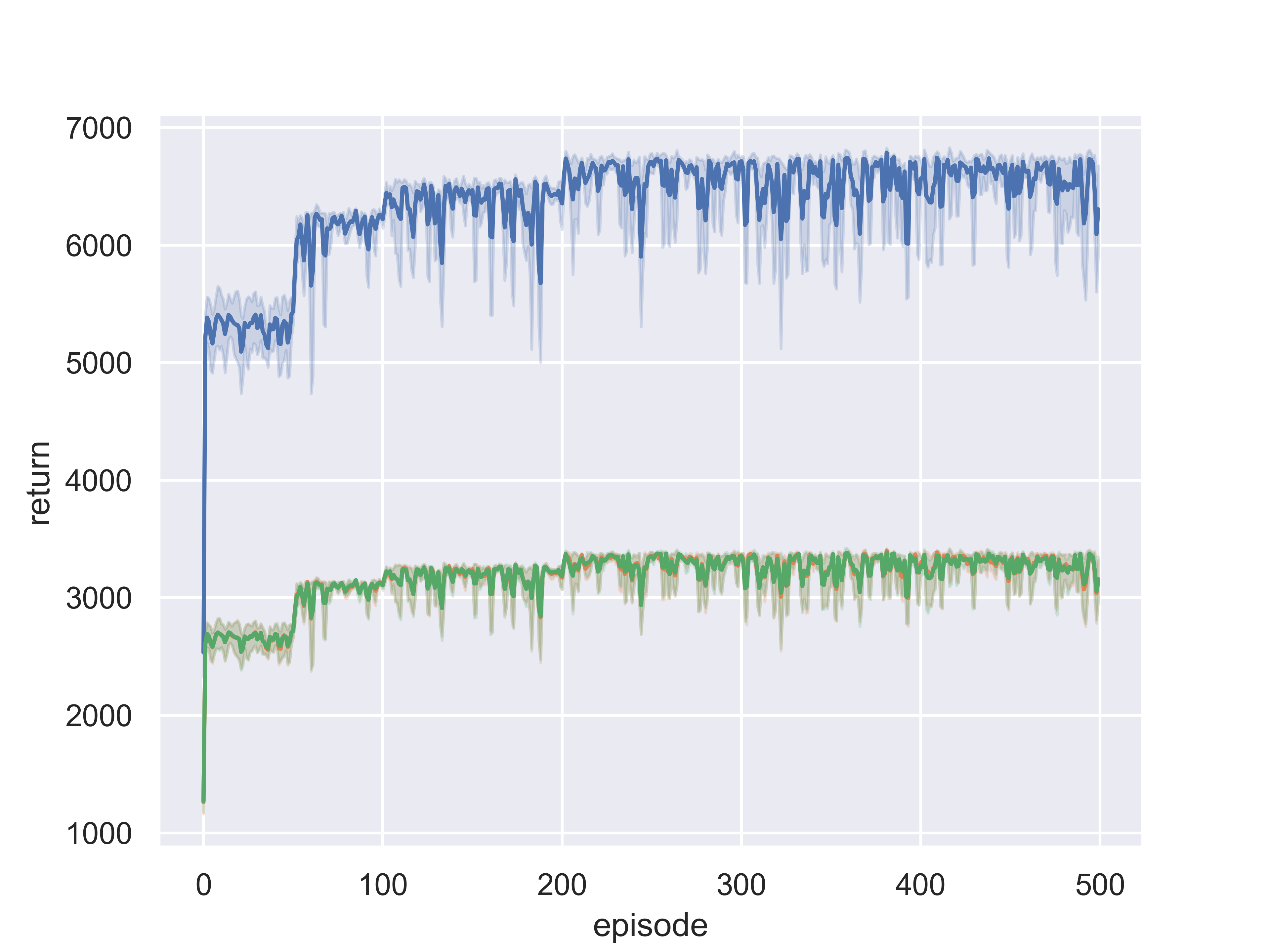}
\caption{Profile of Maximin Q-vectors in Case 2: two arms lifted.}
\label{MM05}
\end{minipage}
\end{figure*}

\subsubsection{Case 2: Unbalanced Action Costs}

The unbalanced cost of action $c = (0, -5)$ is considered in this case, meaning that the elevation action of the left arm incurs no cost, while the same action for the right arm incurs a cost of $-5$. Consequently, the Q-vector variations are modified as shown in Table~\ref{tablecase2}. The optimal actions change for some Q-vectors, and the experimental results are presented in Figures~\ref{Max05} through~\ref{MM05}.

\begin{table}[h]
\centering
\begin{tabular}{|cc|cc|}
\hline
\multicolumn{2}{|c|}{\multirow{2}{*}{\bf Max: $\mathbf{(-0.1, 0)}$}} & \multicolumn{2}{c|}{Right arm}  \\ 
  & & 0 & -0.1 \\
\hline \multirow{2}{*}{Left arm} & 0 & $(0,\; 0)$ &  $(p_1,\; p_1-5)$ \\
& -0.1 & $(p_1,\; \mathbf{p_1})$ &  $(\mathbf{p_2},\; p_2-5)$ \\
\hline
\hline
%%%
\multicolumn{2}{|c|}{\multirow{2}{*}{\bf Nash: $\mathbf{(-0.1, 0)}$ }} & \multicolumn{2}{c|}{Right arm}  \\ 
& & 0 & -0.1 \\
\hline \multirow{2}{*}{Left arm} & 0 & $(0,\; 0)$ &  $(p_1,\; p_1-5)$ \\
& -0.1 & $\mathbf{(p_1,\; p_1)}$ &  $(p_2,\; p_2-5)$ \\
\hline
\hline
%%%
\multicolumn{2}{|c|}{\bf Maximin:} & \multicolumn{2}{c|}{Right arm}  \\ 
\multicolumn{2}{|c|}{ $\mathbf{(-0.1, -0.1)}$} & 0 & -0.1 \\
\hline \multirow{2}{*}{Left arm} & 0 & $(0,\; 0)$ &  $(p_1,\; \mathbf{p_1-5})$ \\
& -0.1 & $(\mathbf{p_1},\; p_1)$ &  $(p_2,\; p_2-5)$ \\
\hline
\hline
\end{tabular}
\vspace{3mm}
\caption{Q-vector variations and optimal actions in Case 2.}
\label{tablecase2}
\end{table}

First, for the Max Q-vector, the optimal action has changed from $(0,0)$ to $(-0.1,0)$. The new optimal action implies that the left arm is lifted while the right arm remains stationary. This behavior was successfully learned and is shown in Figure~\ref{Max05}.
 
Secondly, for the Nash Q-vector, the optimal action has also changed to $(-0.1,0)$. Similarly, it was observed in the experiments that the left arm was predominantly lifted, as shown in Figure~\ref{Nash05}. However, in some experiments as shown in Figure~\ref{Nash05_1}, it was observed that the left arm was lifted first, followed by the right arm. Eventually, both arms worked together to lift the pot. This behavior can be explained as follows.

 When the left arm is lifted alone, the action $(-0.1, 0)$ indicates that the left robotic arm continues to lift, resulting in the right robotic arm not being able to continue to grip the object, and the right side's distance reward decreases, represented by $-\delta$ with $\delta > 0$. The action $(0, -0.1)$ means that the right robotic arm makes a lifting action and the left arm stays, then the tilt angle of the task target decreases, implying an extra reward to both arms, denoted as $+\delta$. With the reward weights set accordingly, $\delta$ is sufficiently large to satisfy $p_1 - \delta < p_2 - 5$ and $p_1 + \delta > p_2$. With these extra modifications in the reward, the Q-vector variations are listed in Table~\ref{tablecase2b}, implying a different Nash action $(0,-0.1)$. Following this action, the left arm remains stationary and the right arm is lifted. They eventually lifted the pot to the highest position.

\begin{table}[h]
\centering
\begin{tabular}{|cc|cc|}
\hline
%%%
\multicolumn{2}{|c|}{\multirow{2}{*}{\bf Nash: $\mathbf{(0, -0.1)}$ }} & \multicolumn{2}{c|}{Right arm}  \\ 
& & 0 & -0.1 \\
\hline \multirow{2}{*}{Left arm} & 0 & $(0,\; 0)$ &   $\mathbf{(p_1+\delta,\; p_1+\delta-5)}$ \\
& -0.1 & $ (p_1,\; p_1-\delta)$ &  $(p_2,\; p_2-5)$ \\
\hline
\hline
\end{tabular}
\vspace{3mm}
\caption{Modified Q-vector variations and the Nash action.}
\label{tablecase2b}
\end{table}

Finally, for the Maximin Q-vector, the optimal action remains unchanged. Similar to Case~1, both arms were simultaneously lifted, raising the object to its highest point, as shown in Figure~\ref{finalposition}. The training performance is presented in Figure~\ref{MM05}, verifying the effectiveness of the learning algorithm.

\section{Conclusion}
\label{sec.Conclusion}

This paper presents a novel DQN algorithm designed to learn Q-vectors using Max, Nash, and Maximin strategies, addressing the complexities of deriving optimal policies in MARL environments. The proposed method effectively handles the varying Q-values resulting from individual agent rewards and demonstrates its efficacy in a dual robotic arm lifting scenario. Future work will focus on scaling the algorithm to larger multi-agent systems, exploring additional game-theoretic strategies, and applying the approach to real-world applications to further validate and enhance its practical utility.

\bibliographystyle{abbrv}
\bibliography{sample}

\end{document}